\pdfoutput=1

\documentclass[11pt]{article}

\usepackage{ACL2023}

\usepackage{booktabs} 
\usepackage{graphicx}
\usepackage{multirow}

\usepackage{xcolor}
\usepackage[linesnumbered,ruled,vlined]{algorithm2e}

\SetCommentSty{mycommfont}
\SetKwInput{KwInput}{Input} 
\SetKwInput{KwOutput}{Output} 

\newcommand{\topic}[1]{\textsc{#1}} 

\usepackage{times}
\usepackage{latexsym}

\usepackage[T1]{fontenc}

\usepackage[utf8]{inputenc}

\usepackage{microtype}

\usepackage{inconsolata}

\usepackage{enumitem}

\usepackage{xcolor}
\usepackage[normalem]{ulem}

\title{A Shocking Amount of the Web is Machine Translated: \\Insights from Multi-Way Parallelism}

\author{Brian Thompson,\thanks{\hspace{1.5mm}Corresponding author}$\hspace{1.9mm}^{1}$\hspace{0.1mm} Mehak Preet Dhaliwal,\thanks{\hspace{1.5mm}Work conducted during an internship at Amazon.}\hspace{1.9mm}$^{2}$\hspace{0.1mm} Peter Frisch,$^{1}$ Tobias Domhan,$^{3}$ Marcello Federico$^{1}$ \\
$^{1}$AWS AI Labs \hspace{1.0mm} $^{2}$UC Santa Barbara \hspace{1.0mm} $^{3}$Amazon \\
\href{mailto:brianjt@amazon.com}{brianjt@amazon.com}}

\begin{document}
\maketitle
\begin{abstract}

We show that content on the web is often translated into many languages, and the low quality of these multi-way translations indicates they were likely created using Machine Translation (MT).
Multi-way parallel, machine generated content 
not only dominates the translations in lower resource languages;
it also constitutes a large fraction of the total web content in those languages.
We also find evidence of a selection bias in the type of content which is translated into many languages,
consistent with low quality English content being translated en masse into many lower resource languages, via MT.
Our work raises serious concerns about training models such as multilingual large language models on both monolingual and bilingual data scraped from the web.

\end{abstract}

\section{Introduction}

Modern AI is enabled by huge amounts of training data, typically several hundred billion tokens to a few trillion tokens \cite{sun2021ernie, chowdhery2022palm, touvron2023llama, almazrouei2023falcon}.
Training at this scale is only possible with web-scraped data.

\begin{figure}[!t]
    \centering
    \includegraphics[width=6.0cm]{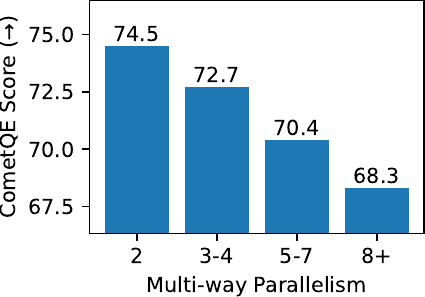}
    \caption{The more languages a sentence has been translated into (``Multi-way Parallelism''), the lower quality the translations are, suggesting a higher prevalence of machine translation.
    See \autoref{analysis:bitext} for more details.}
    \label{fig:fig1}
\end{figure}

We explore the effects that the long-term availability of low cost Machine Translation (MT) has had on the web.\footnote{Free MT has been available online since late 1997 \cite{gaspari-hutchins-2007-online}, around the same time that MT researchers began scraping the web for training data \cite{resnik-1998-parallel}, and commercial systems have been available since the 1970s \cite{hutchins1995machine}.}
We show that content on the web is often translated into many languages, and the quality of these multi-way translations indicates they were primarily created using MT: see \autoref{fig:fig1}. 
Machine generated, multi-way parallel translations not only dominate the total amount of translated content on the web in lower resource languages where MT is available,
it also constitutes a \emph{large fraction of the total web content} in those languages. 
We also find evidence of a selection bias in the \emph{type} of content which is translated into many languages, and therefore over represented in lower resource languages: This content is shorter, more predictable, and has a different topic distribution compared to content translated into a single language. 
A limited investigation suggests this selection bias is the result of low quality content generated in English (likely produced to generate ad revenue) and translated en masse into many lower resource languages via MT (again, likely to generate ad revenue).

Our findings raise numerous concerns for multilingual model builders: 
Fluency (especially across sentences) and accuracy are lower for MT data,\footnote{MT technology has improved dramatically over the last decade, but still falls short of human quality \cite{freitag-etal-2023-results}. MT content has been added to the web over many years using MT systems available at the time, so much of the MT on the web is likely very low quality by modern standards.}
which could produce less fluent models with more hallucinations, and the selection bias indicates the data may be of lower quality, even before considering MT errors.
Data quality is crucial in Large Language Model (LLM) training, where high quality corpora like books and Wikipedia articles are typically upsampled several times \cite{brown-etal-2020-language, gao2020pile, rae2021scaling, le-scao-etal-2022-language}. 

Our findings also help to explain why low-resource MT \cite{khan2017machine, duh18multitarget, nllbteam2022language} is challenging, and why filtering noise  \cite{khayrallah-koehn-2018-impact} from web-scraped bitext \cite{junczys-dowmunt-2018-dual, chaudhary-etal-2019-low} is beneficial for MT training \cite{koehn-etal-2018-findings, koehn-etal-2019-findings, koehn-etal-2020-findings, sloto-etal-2023-findings}.

To enable analysis, we create the largest multi-way corpus to date, consisting of 6.4B unique sentences in 90 languages. We release code to reproduce our
corpus and analysis.\footnote{\url{https://github.com/amazon-science/multi-way-parallel-ccmatrix/}. Corpus creation has been optimized to run in about one day on a single i4i.32xlarge AWS instance.}

\section{Related Work}

Our work is inspired by several recent efforts which seek to understand the characteristics of large scale corpora 
\cite{mehmood-shafiq-2017-understanding, dodge-etal-2021-documenting, kreutzer-etal-2022-quality, brannon-etal-2023-dubbing}.
Many works have detected machine translation \cite{kurokawa-etal-2009-automatic,arase-zhou-2013-machine,aharoni-etal-2014-automatic},
but we are not aware of prior work using multi-way parallelism to do so.
\citet{freitag-firat-2020-complete} explored multi-way parallelism
with the goal of improving multilingual MT.

Exploring multi-way parallelism on the web requires a curated representation of translated content from the web. 
We build upon ccMatrix \cite{schwenk-etal-2021-ccmatrix}, which is in turn based on Common Crawl.\footnote{\url{https://commoncrawl.org/}}
Common Crawl is a long running web-scraping project which maintains a free, open source repository of web-scraped data.
ccMatrix is created by embedding Common Crawl sentences 
into a multilingual space using LASER \cite{artetxe-schwenk-2019-massively} 
and then finding bilingual translation pairs using fast approximate nearest neighbor search \cite{johnson2019billion}. 
We choose ccMatrix over a corpus from a traditional bitext mining process of document alignment \cite{resnik-smith-2003-web, buck-koehn-2016-quick, thompson-koehn-2020-exploiting} followed by sentence alignment \cite{gale-church-1993-program, sennrich-volk-2010-mt, thompson-koehn-2019-vecalign}
because it is the largest corpus available at the time of writing (in both number of sentences and language coverage).

\section{Corpus Creation: MWccMatrix}\label{mwccmatrix}

We create a multi-way parallel representation of the web,
consisting of translation \emph{tuples} containing \emph{two or more} sentences in different languages which are translations of each other.\footnote{Unless otherwise noted, we use the term ``translation'' to mean a sentence which appears in a translation tuple -- i.e.\ we do not attempt to distinguish whether that sentence was translated into or out of a given language.}
As a trivial example, (``hello'', ``hola'') in English-Spanish and (``hello'', ``olá'') in English-Portuguese combine to make (En:``hello'', Es:``hola'', Pt:``olá'').
We denote this corpus Multi-Way ccMatrix (MWccMatrix).

We iterate through all bitext in ccMatrix, from highest to lowest LASER margin score, adding sentence pairs as new tuples in MWccMatrix when neither sentence is already in the new corpus, and expanding tuples already in the new corpus when 
one sentence or the other (but not both) 
is already present. 
This deduplicates the corpus (i.e.\ adds each unique sentence only once),
but allows for more than one sentence in the same language to be added to a given tuple,
which tend to differ primarily in punctuation/capitalization (i.e.\ near duplicates). 
Therefore, we remove all but the first sentence added to each tuple in a given language.
Deduplication across language pairs brings the total number of sentences down from 21.7B total sentences (10.9B sentence pairs) to 7.9B unique sentences in 2.2B tuples,
and near duplicate removal brings it down to 6.4B.
Pseudocode and a description of the optimizations required to make corpus creation tractable are provided in \autoref{appendix:alg}.

\section{Analysis}\label{analysis}

\begin{figure}[!ht]
    \centering
    \includegraphics[width=6.8cm]{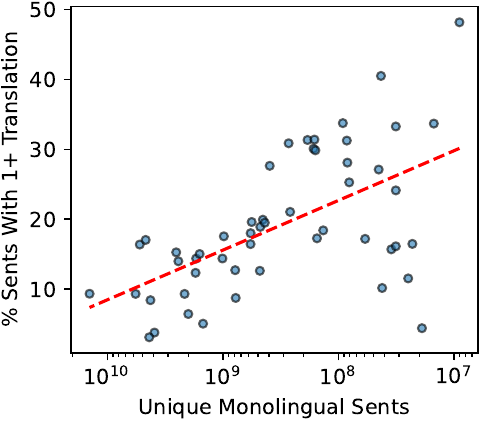}

    \caption{Fraction of the total monolingual data used to create ccMatrix with one or more translation, in the 54 languages for which we can compute it. 
    See \autoref{appendix:scatterplot} for a larger plot with language codes.
    }
    \label{fig:percent_parallel}
\end{figure}

\subsection{Much of the Web is Translated}\label{analysis:lotsoftranslations}

We compared the total number of unique sentences (before removing near-duplicates) in MWccMatrix to the total number of unique sentences from the Common Crawl snapshots that the data is based on, as reported by \citet{schwenk-etal-2021-ccmatrix}.
They only report the number of unique sentences for the 54 (of 90) largest resource languages,
so we cannot compute the fraction of sentences with one or more translations in the 36 lowest-resource languages.
The percentage of unique monolingual sentences which have at least one translation is quite high, even for some high resource languages (e.g.\ 9.4\% of English, 17.5\% of French): see \autoref{fig:percent_parallel}.

\begin{table}[!t]
    \centering
    \small
    \begin{tabular}{l r r r r}
    \toprule
Parallelism & \# tuples  & \% tuples & \# sents  & \% sents \\
\midrule
2   & 1,368   &  62.5\%  & 2,736  &  42.9\%  \\
3-4 &   573   &  26.2\%  & 1,895  &  29.7\%  \\
5-7 &   177   &   8.1\%  & 1,004  &  15.7\%  \\
8+  &    70   &   3.2\%  &   745  &  11.7\%  \\
\midrule
Total   & 2,188 & 100.0\% &  6,379 &  100.0\% \\
\bottomrule
    \end{tabular}
    \caption{MWccMatrix statistics. Numbers in millions. 37.5\% of tuples are multi-way parallel, but 57.1\% of all sentences come from multi-way parallel tuples. 
    }
    \label{tab:basic_stats}
\end{table}

\subsection{Translations on the Web are Highly Multi-way Parallel}\label{analysis:mwp}

\begin{figure*}[!ht]
    \centering
    \includegraphics[width=16cm]{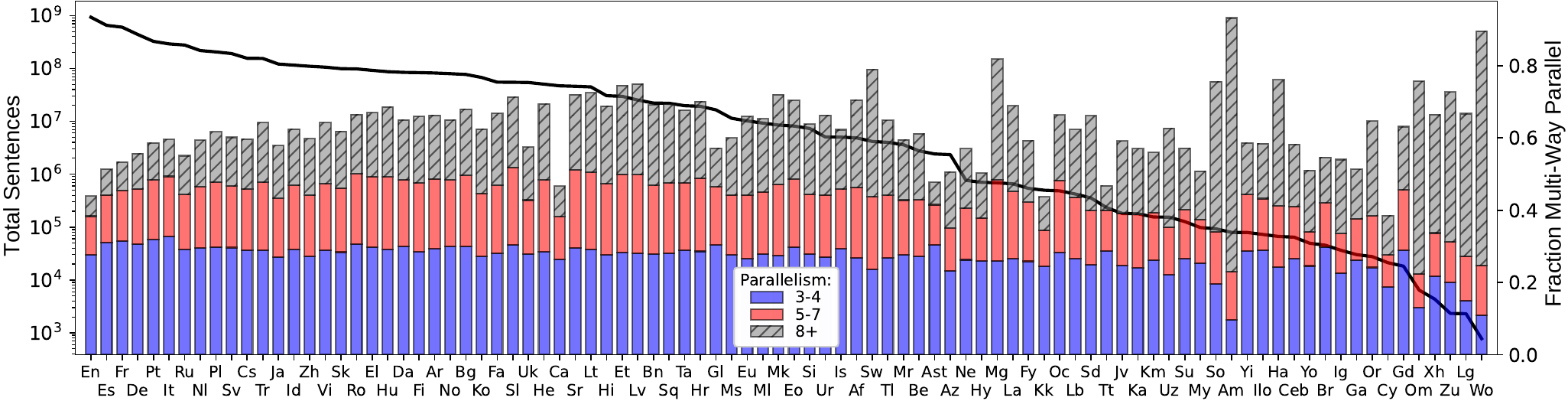}
    \caption{Fraction of parallel data in each language which is multi-way parallel (bar chart, right y-axis) and number of unique sentences (solid black line, left y-axis, log scale) by language (x-axis). 
    Low-resource languages exhibit a dramatic increase in the amount of highly multi-way parallel data (hatched gray bars).
    }
    \label{fig:barplot}
\end{figure*}

Of the 6.38B sentences in our 2.19B translation tuples, 3.63B (57.1\%) are in multi-way parallel\footnote{We use ``multi-way parallelism'' (or simply ``parallelism'') to refer to the size of the translation tuple that that sentence is in. For example, a sentence with parallelism of 5 comes from a tuple of size 5, which contains the given sentence plus translations in 4 other languages.}  (3+ languages) tuples: see \autoref{tab:basic_stats}. 
Lower resource languages tend to be more multi-way parallel, with 
the 10 highest-resourced languages in ccMatrix having an average parallelism of 4.0,
and the 10 lowest-resource languages in ccMatrix having an average parallelism of 8.6 (see \autoref{appendix:barplot} for all languages), and this increase is driven by an increase in highly multi-way parallel (8+) sentences: see \autoref{fig:barplot}.

\subsection{Multi-way Parallel Data is Shorter and Simpler}\label{analysis:monolingual}

\begin{table} %
    \centering
    \small
    \begin{tabular}{l r r r r r}
\toprule
Parallelism    &  De  &   En  & Fr   & Ja    & Zh   \\
\midrule    
2   & \textbf{95.2} & \textbf{103.7} & \textbf{96.8} & \textbf{25.2}  & \textbf{27.8} \\
3-4 & 90.4 &  86.0 & 88.7 & 23.6  & 24.8 \\
5-7 & 82.6 &  71.2 & 80.5 & 23.2  & 22.8 \\
8+  & 71.6 &  59.9 & 70.0 & 22.3  & 19.1 \\
\bottomrule
    \end{tabular}
    \caption{Sentence length (in characters) as a function of multi-way parallelism, in several languages. Multi-way parallelism is associated with shorter content.}
    \label{fig:sent_len}
\end{table}

\begin{figure}
    \centering
    \includegraphics[width=6.0cm]{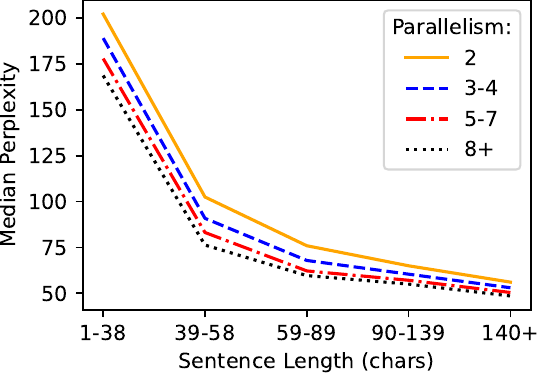}
    \caption{Median perplexity (measured by GPT-2) vs multi-way parallelism, in English. We stratify by sentence length, as shorter content tends to have higher perplexity, likely due to GPT-2 having no or little context for predicting the first few words. More multi-way parallel data has lower perplexity (i.e. easier to predict).}
    \label{fig:ppl}
\end{figure}

We perform monolingual analysis 
to explore how data varies with multi-way parallelism.
We find that more multi-way parallel sentences are shorter in length (see \autoref{fig:sent_len}) and have lower perplexity (i.e.\ are easier to predict) as measured by GPT-2 \cite{radford2019language}: see \autoref{fig:ppl}.

\subsection{Multi-way Parallel Translations are Lower Quality}\label{analysis:bitext}

\begin{table*}[!h]
    \centering
    \small
\begin{tabular}{l|cccccccc|l}
\toprule
Parallelism & En$\rightarrow$De & De$\rightarrow$En & Fr$\rightarrow$De & De$\rightarrow$Fr & En$\rightarrow$Ja & Ja$\rightarrow$En & En$\rightarrow$Zh & Zh$\rightarrow$En & \multicolumn{1}{c}{AVG} \\
\midrule
\textbf{2} & \textbf{76.5} & \textbf{76.1} & \textbf{73.3} & \textbf{74.6} & \textbf{73.6} & \textbf{71.9} & \textbf{74.8} & \textbf{75.4} & \textbf{74.5} \\
3-4 & 74.3 & 74.2 & 72.3 & 73.7 & 72.0 & 70.6 & 72.1 & 72.5 & 72.7 (-1.8)\\
5-7 & 71.9 & 71.8 & 70.0 & 71.3 & 70.5 & 69.2 & 69.1 & 69.5 & 70.4 (-4.1) \\
8+  & 69.7 & 69.8 & 67.5 & 68.6 & 69.7 & 68.6 & 66.1 & 66.6 & 68.3 (-6.2) \\
\bottomrule
\end{tabular}
    \caption{Bitext quality (as measured by CometQE) as a function of multi-way parallelism,
    for random 1M subsets in various language pairs. Multi-way parallel translations are lower quality. Average scores are visualized in \autoref{fig:fig1}.}
    \label{tab:cometKiwi}
\end{table*}

\begin{figure}
    \centering
    \includegraphics[width=6.0cm]{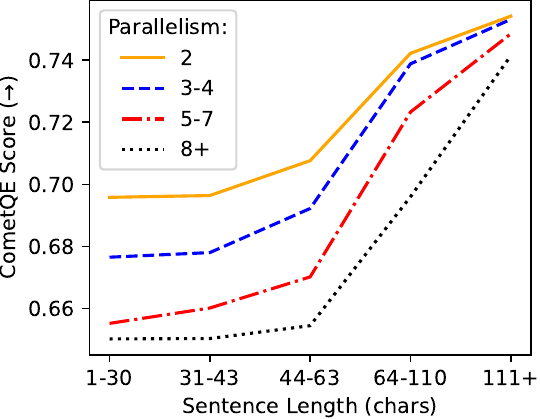}
    \caption{CometQE score vs sentence length (average of source and target, in characters), for Fr-De. Other language pairs (not shown) are very similar. Quality differences between levels of multi-way parallelism holds across sentence length.}
    \label{fig:kiwi_vs_len}
\end{figure}

\begin{figure}
    \centering
    \includegraphics[width=6.0cm]{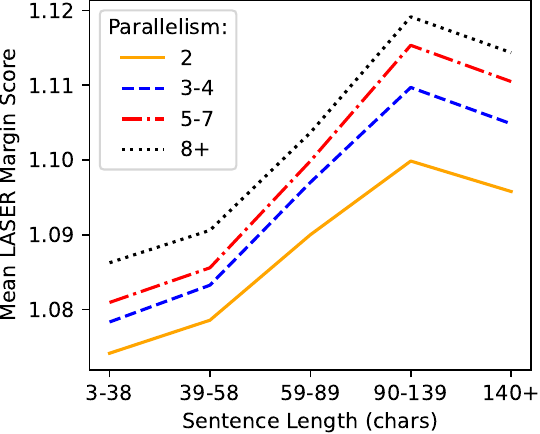}
    \caption{LASER margin scores as a function of multi-way parallelism and sentence length, in English. Trends in other languages that we investigated (French, German, Chinese, Japanese) were very similar (not shown).}
    \label{fig:marginscores}
\end{figure}

We evaluate the quality of translations on the web using Quality Estimation (QE), with the CometQE model \cite{rei-etal-2022-cometkiwi}, across different levels of multi-way parallelism.
Modern quality estimation methods are nearly on par with reference-based metrics \cite{freitag-etal-2023-results} and have been shown to perform well on noisy web data \cite{peter-etal-2023-theres}. 
As QE does not require human annotation or human references, it allows us to evaluate a very large data sample (1M samples per language pair) 
and many language pairs.\footnote{We select from WMT language pairs as CometQE is trained on WMT annotations, thus we expect CometQE to be most accurate in those language pairs.}

We find that highly multi-way parallel translations are significantly lower quality (6.2 CometQE points worse) than 2-way parallel translations.
This trend is consistent across all 8 language pair directions we considered: see \autoref{tab:cometKiwi}.
Since length could interact with cometQE scores, we verified that these results also hold across sentence length: see \autoref{fig:kiwi_vs_len}.

\begin{table}
    \centering
    \small
    \begin{tabular}{l l l l}
\toprule
Langs  & Ref  & MT  & $\Delta$ \\
\midrule
En$\rightarrow$De & 86.7\% & 89.5\% +/- 0.6\%   &  2.8\% \\
De$\rightarrow$En & 87.4\% & 90.1\% +/- 0.4\%   &  2.8\% \\
Zh$\rightarrow$En & 82.3\% & 85.3\% +/- 0.3\%   &  3.0\% \\
En$\rightarrow$Zh & 82.6\% & 85.8\% +/- 0.4\%   &  3.1\% \\
En$\rightarrow$Ja & 78.8\% & 82.7\% +/- 1.1\%   &  3.9\% \\
De$\rightarrow$Fr & 87.4\% & 90.1\% +/- 0.4\%   &  2.8\% \\
Fr$\rightarrow$De & 90.5\% & 91.4\% +/- 0.6\%   &  0.9\% \\
\bottomrule
    \end{tabular}
    \caption{LASER cosine similarity between source and human reference (``Ref'') vs mean and standard deviation for Online-Y, Online-G, Online-A, Online-W, and Online-B (``MT'') from WMT 2022. In cases where there is more than one human reference, we average the cosine similarities. We find that LASER has a bias for MT output, of about 2.8\% on average. Note that Ja$\rightarrow$En was not included among the WMT language pairs.}
    \label{fig:laser_likes_mt}
\end{table}

We also investigate how LASER margin score varies with multi-way parallelism.
Multi-way parallel data tends to have higher margin scores: see \autoref{fig:marginscores}. 
Further investigation reveals that LASER has a strong bias for MT output over human translations (see \autoref{fig:laser_likes_mt}),
thus LASER margin scores for more multi-way parallel content are consistent with multi-way parallel data being MT.
LASER's preference for MT is likely because LASER is based on a small MT model. Similar phenomenon has been observed \cite{freitag-etal-2021-results} in the Prism metric \cite{thompson-post-2020-automatic, thompson-post-2020-paraphrase}, which is also based on an MT model.

\begin{table}[!ht]
\centering
\small
\addtolength{\tabcolsep}{-1pt}
\begin{tabular}{lrrrr}
\toprule
       & \multicolumn{4}{c}{Parallelism} \\
Topic & \multicolumn{1}{c}{2} & \multicolumn{1}{c}{3-4} & \multicolumn{1}{c}{5-7} & \multicolumn{1}{c}{8+} \\
\midrule
Autos/Vehicles/Transit    &  2.2 &  2.0 &  1.6 &  1.6 \\
Beauty/Fitness/Health.    &  6.5 &  7.2 &  6.4 &  6.4 \\
Books/Arts/Entertainment  &  8.0 &  7.3 &  6.3 &  5.4 \\
Business/Industry/Finance & 10.6 &  7.6 &  6.0 &  6.6 \\
Computers/Electronics     &  3.7 &  3.4 &  4.5 &  3.5 \\
Conversation/Opinion      & 22.5 & 29.9 & 33.3 & 40.1 \\
Food/Drink                &  1.7 &  1.8 &  1.7 &  0.9 \\
Hobbies/Leisure           &  0.9 &  0.9 &  1.4 &  1.0 \\
Home/Garden               &  0.9 &  0.7 &  0.9 &  0.5 \\
Internet/Telecom          &  3.2 &  3.0 &  2.1 &  2.6 \\
Jobs/Education            &  6.6 &  4.7 &  5.0 &  5.0 \\
Law/Government            &  7.2 &  6.0 &  5.2 &  4.8 \\
News/Information          & 12.2 & 11.8 & 10.5 &  9.5 \\
Pets/Animals              &  1.2 &  1.4 &  1.2 &  1.0 \\
Real Estate               &  1.3 &  0.8 &  0.5 &  0.5 \\
Religion                  &  3.4 &  4.7 &  5.6 &  5.4 \\
Shopping                  &  1.1 &  1.2 &  1.2 &  0.7 \\
Sports/Games              &  3.2 &  2.9 &  3.7 &  2.2 \\
Travel                    &  2.9 &  2.6 &  2.5 &  1.9 \\
Other/Cannot tell         &  0.2 &  0.1 &  0.5 &  0.7 \\
\bottomrule
\end{tabular}\caption{Percentage of corpus which human annotators classified as each topic, for various levels of parallelism. 
}\label{table:topics}
\end{table}

\subsection{Multi-way Parallel Data has Topical Bias}\label{analysis:topic}

We had professional linguists classify\footnote{Full annotator guidelines are provided in \autoref{appendix:guidelines}} 10,000 randomly selected English sentences as one of the 20 topics given in \autoref{table:topics},
based on the high-level Topics API categories.\footnote{\url{https://cloud.google.com/natural-language/docs/categories}}
We observe a dramatic shift in the distribution of topics when comparing 2-way to 8+ way parallel data, 
with  \topic{Conversation \& Opinion} increasing from 22.5\% to 40.1\%. 

We manually inspected a random sample of 100 highly multi-way parallel sentences from the \topic{Conversation \& Opinion} topic and found them hard to characterize due to the isolated sentences being very short (typically 5-10 words). 
However, searching the web for the sentences was enlightening: the vast majority came from articles that we characterized as low quality, 
requiring little or no expertise or advance effort to create,
on topics like being taken more seriously at work, being careful about your choices, six tips for new boat owners, deciding to be happy, etc. 
Furthermore, we were unable to find any translationese or other errors that would suggest the articles were being translated into English (either by human translators or MT),
suggesting it is instead being generated in English and translated to other languages.

\section{Discussion \& Conclusion}

Experiments with QE (see \autoref{analysis:bitext}) strongly suggest that highly multi-way parallel translations are generated by MT.
In lower resource languages, \emph{most} translations are multi-way parallel (see \autoref{analysis:mwp}), suggesting that MT content dominates translation content.
Furthermore, a large fraction of the \emph{total} sentences in lower resource languages have at least one translation (see \autoref{analysis:lotsoftranslations}),
implying that a large fraction of the \emph{total web} in those languages is MT generated. 

Several observations point to a selection bias in the \emph{type} of data which is translated into many languages, compared to data translated into a single language: it is shorter and more predictable (see \autoref{analysis:monolingual}),
and substantially more likely to be from the \topic{Conversation \& Opinion} topic (see \autoref{analysis:topic}). 
Since translations of this data constitute a substantial portion of the total data in low-resource languages, this bias will also appear in low resource languages. 

An investigation into the increase in \topic{Conversation \& Opinion} data suggests that this selection bias is the result of low quality content (likely produced to generate ad revenue) being translated via MT en masse into many lower resource languages (again likely for the purpose of generating ad revenue). It also suggests that such data originates in English and is translated into other languages. Additional investigation would be required to understand if this finding generalizes to other topics, languages, and levels of multi-way parallelism.

Our findings also suggest some ways to address the problem of MT output in web-scraped training data:
MT detection, which has typically been proposed to filter bitext, could also help to filter monolingual text in lower resource languages. 
It also suggests that multi-way parallelism
is a promising way to detect low quality, machine translated data, especially in lower resource languages, to filter both bilingual and monolingual data. 

\section*{Limitations}\label{limitations}

Our study covers 90 of the most common languages on the web, where MT tends to be available. We would obviously not expect the trends we observe regarding low resource languages to extend into the long tail of low-resource languages that are not currently supported by MT.

All our analysis is performed at the sentence level, because ccMatrix is at the sentence level; this makes some analysis (e.g.\ topic analysis) difficult and/or ambiguous. 
We would have preferred to use a corpus which is aligned 
at the document level to enable document level analysis and evaluation \cite{laubli-etal-2018-machine, toral-etal-2018-attaining,  vernikos-etal-2022-embarrassingly}, 
but no such corpus is publicly available. 

Similarly, our analysis does not take advantage of any cues that might be present in a web page to indicate its content is MT generated. 
However, in personal communications with the authors of Paracrawl \cite{banon-etal-2020-paracrawl}, they note that in low-resource languages supported by popular MT systems, simple rules\footnote{\url{https://github.com/paracrawl/cirrus-scripts/blob/master/mt-filter-list.annotated}} to remove data from common translation plug-ins filter out most of their scraped bitext. This observation is consistent with our findings.

We use CometQE to evaluate translation quality. CometQE is trained on human annotations of translation quality from many years of WMT \cite{kocmi-etal-2023-findings} evaluations. The web data in our experiments likely has a very different domain distribution than WMT data, and trained metrics like CometQE have been shown to exhibit a performance drop when moving from WMT to other domains \cite{zouhar2024finetuned}.

Our analysis of the web
is based on bitext mined from the web.
As such, shortcomings or biases in web scraping and bitext mining could affect our results. 
Common Crawl provides only a sample of the web, and 
biases have been demonstrated in web scraping \cite{mehmood-shafiq-2017-understanding, dodge-etal-2021-documenting}.
Common Crawl follows links within web pages to find new pages, and web pages sometimes have links to translations of the same page in another language, so Common Crawl may be more likely to find web pages which are translations of pages it has already found than other, new pages. This should be mitigated at least in part by combining many Common Crawl snapshots, as is done in ccMatrix. 

The 32 snapshots of Common Crawl used in ccMatrix were collected between December 2017 to February 2020 \cite{schwenk-etal-2021-ccmatrix}. We are not aware of a more recent, publicly available  corpus that would enable this kind of analysis. 

The ccMatrix corpus creation process relies on LASER margin scores (as does our process to create MWccMatrix). LASER is known to have lower recall in lower-resource languages \cite{feng-etal-2022-language} and as we show in this work (\autoref{fig:laser_likes_mt}), has a preference for MT over human translations. ccMatrix also uses approximate nearest neighbor search \cite{johnson2019billion}, which trades off some recall performance in order to make searches computationally feasible. 

Our analysis by language / language pair relies on automatic language identification (LID). Shortcomings have also been noted in automatic LID, especially in low-resource languages \cite{caswell-etal-2020-language, kreutzer-etal-2022-quality}.

\section*{Acknowledgements}

We would like to thank 
Mohaddeseh Bastan,
Anna Currey,
Kenneth Heafield,
Huda Khayrallah,
Hieu Hoang,
Surafel Lakew,
Prashant Mathur,
Yogesh Virkar, 
and the anonymous ACL reviewers 
for their constructive feedback at various stages of drafting. 
We would also like to think Tatyana Badeka and Jenyuan Wang for assistance with the topic analysis.

\bibliography{main.bbl}
\bibliographystyle{acl_natbib}

\clearpage
\appendix

\section{MWccMatrix Creation: Additional Details}\label{appendix:alg}

\begin{algorithm*}
\caption{Algorithm (simplified for comprehension) used to create multi-way parallel corpus.}\label{algo}
\DontPrintSemicolon
$sent2row  \leftarrow defaultdict(dict)$\;
$sort(bitext, key=marginScore, descending=True)$\;
$numRows\leftarrow 0$\;
\For{srcTxt, srcLang, tgtText, tgtLang, marginScore in bitext}{
    \If{srcTxt not in sent2row[srcLang] and tgtText not in sent2row[tgtLang]}{
        \nl\tcc{add new sentence pair}
        $sent2row[srcLang][srcTxt] \leftarrow (numRows, marginScore)$ \;
        $sent2row[tgtLang][tgtTxt] \leftarrow (numRows, marginScore)$ \;
        $numRows\leftarrow numRows + 1$\;
    }
    \ElseIf{ srcTxt in sent2row[srcLang] } {
        \nl\tcc{srcText in table, join on it}
        $srcRow \leftarrow sent2row[srcLang][srcTxt]$\;
        $sent2row[tgtLang][tgtTxt] \leftarrow (srcRow, marginScore)$ \;
    }
    \ElseIf{ tgtTxt in sent2row[tgtLang] }{
        \nl\tcc{tgtText in table, join on it}
        $tgtRow \leftarrow sent2row[tgtLang][tgtTxt]$\;
         $sent2row[srcLang][srcTxt] \leftarrow (tgtRow, marginScore)$ \;
    }
    \nl\tcc{else both sentences already in table (with higher marginScore), do nothing}
}
\nl\tcc{Invert sent2row}
$row2sent \leftarrow defaultdict(dict)$ \;
\For{lang in langs}{
    \For{sent, (row, marginScore) in sent2row[lang].items()}{
        \If{ row in row2sent[lang]}{
            $\_, oldMarginScore \leftarrow row2sent[lang][row]$
        }
        \Else{
            $oldMarginScore \leftarrow -1$
        }
        \nl\tcc{When we find duplicates/paraphrases, keep the sentence with the highest score}
        \If{marginScore > oldMarginScore}{
            $row2sent[lang][row] \leftarrow (sent, score)$
        }
    }
}
\nl\tcc{Coalesce output translation tuples}
$output \leftarrow []$ \;
\For{row in range(numRows)}{
    $translations \leftarrow dict()$ \;
    \For{lang in langs}{
        \If{row in row2sent[lang]}{
            $translations[lang] \leftarrow row2sent[lang][row]$ \;
        }
    }
    $output.append(translations)$ \;
}
\end{algorithm*}

A simplified version of the algorithm used to create MWccMatrix is provided in \autoref{algo}.

In practice, several optimizations were required to make the process tractable.
Instead of attempting to sort 10.9B sentence pairs by margin score, we approximate the search by binning margin scores and sorting the data into buckets corresponding to the (binned) margin scores, similar to a radix sort. 
The sentences are too large to fit in memory, so we represent the sentences as 64 bit hashes.
Additionally, our scripts are written in python but we use the cykhash\footnote{\url{https://github.com/realead/cykhash}} package, which provides a native C int64 to int64 hashmap. 
Conversion from hashes back to sentences is done in small shards, and the hash$\rightarrow$sent mappings required to reconstruct the data are sharded such that only the mappings required for one shard are loaded in memory at one time. 
Finally, we make extensive use of parallelization (e.g.\ computing margin score bins, sharding data by margin score bin, hashing sentence pairs, etc).

\section{Larger Version of \autoref{fig:percent_parallel}}\label{appendix:scatterplot}

\begin{figure*}[]
    \centering
    \includegraphics[width=16cm]{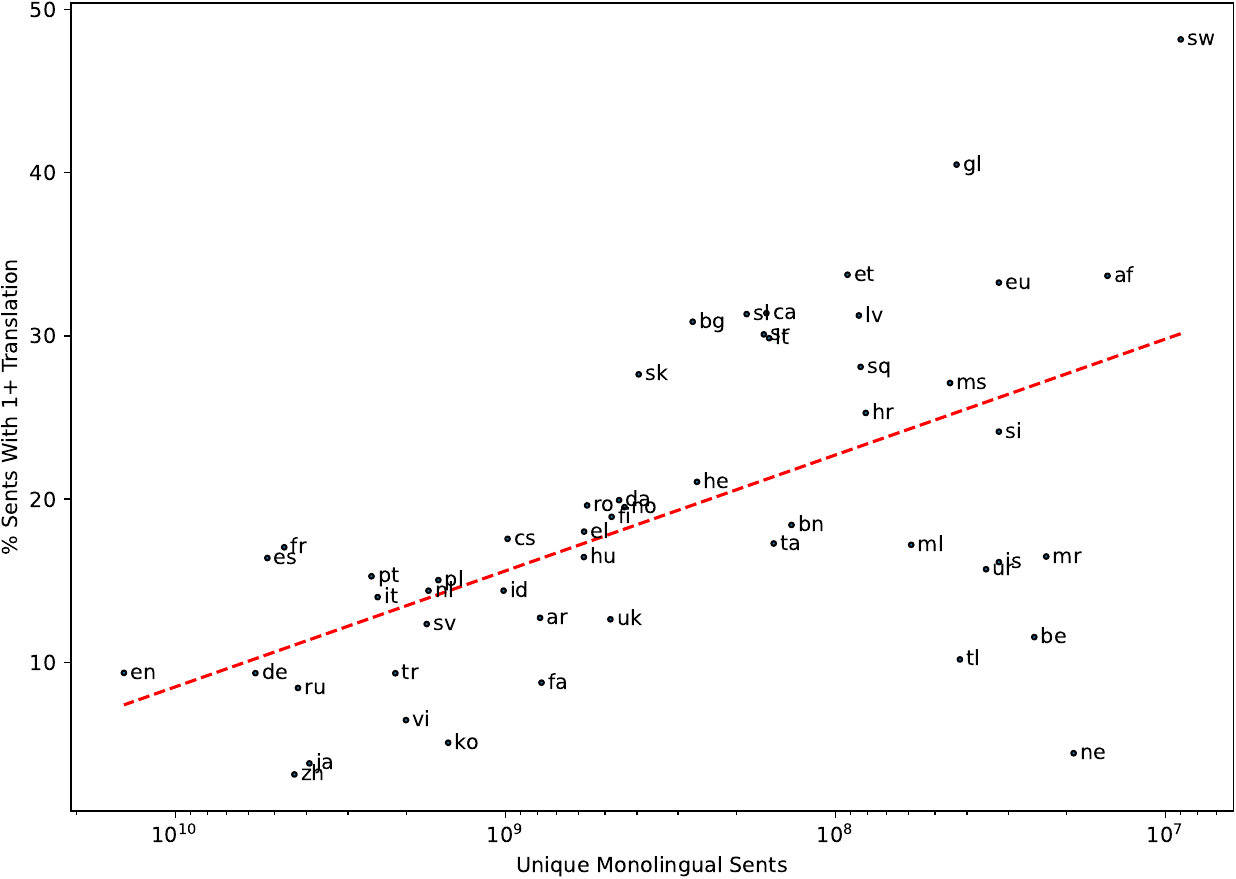}
    \caption{Percentage of unique monolingual sentences which have at least one translation, in each language for which we have the data to compute it.}
    \label{fig:percent_parallel_large}
\end{figure*}

A larger version of \autoref{fig:percent_parallel}, which includes language codes for each language, is provided in \autoref{fig:percent_parallel_large}.
As previously noted, we only have total data sizes for the 54 highest-resource languages, as that is what was reported by \citet{schwenk-etal-2021-ccmatrix}, so we cannot compute this percentage for the 36 lowest-resource languages used in this study.

\section{Multi-way Parallelism by Language}\label{appendix:barplot}

\begin{figure*}[]
    \centering
    \includegraphics[width=16cm]{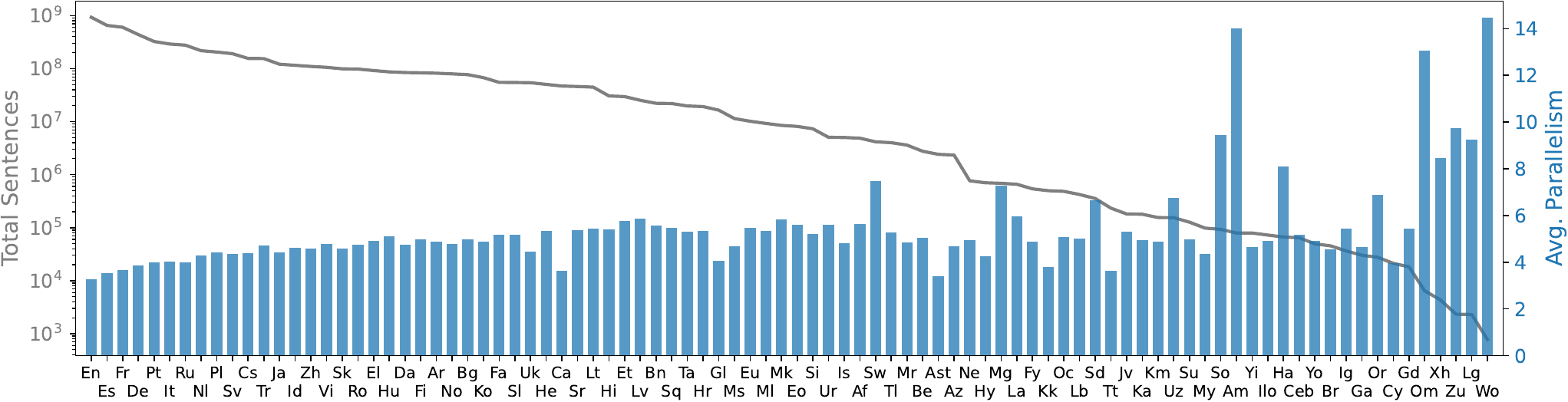}
    \caption{Average multi-way parallelism (blue bars, right y-axis) and number of unique sentences (gray line, left y-axis, log scale) by language (x-axis). Lower-resource languages tend to be more multi-way parallel.}
    \label{fig:basic_figure}
\end{figure*}
 
Average parallelism for each language is shown in \autoref{fig:basic_figure}.

\clearpage
\section{Topic Analysis Annotation Guidelines}\label{appendix:guidelines}

Task: Please identify the most relevant topic for each sentence using the pre-defined list. Assign the correct label to each sentence. Make sure to familiarize yourself with the list before working on the task.\\

\begin{tabular}{ll}
\toprule
Index & Topic \\\midrule
1 & Autos/Vehicles/Transit    \\
2 & Beauty/Fitness/Health.    \\
3 & Books/Arts/Entertainment  \\
4 & Business/Industry/Finance \\
5 & Computers/Electronics     \\
6 & Conversation/Opinion      \\
7 & Food/Drink                \\
8 & Hobbies/Leisure           \\
9 & Home/Garden               \\
10 & Internet/Telecom          \\
11 & Jobs/Education            \\
12 & Law/Government            \\
13 & News/Information          \\
14 & Pets/Animals              \\
15 & Real Estate               \\
16 & Religion                  \\
17 & Shopping                  \\
18 & Sports/Games              \\
19 & Travel                    \\
20 & Other/Cannot tell         \\
\bottomrule
\end{tabular}

\vspace{0.7cm}

\noindent Here is the list of sub-topics that may help you develop a better idea of what belongs to each of these topics: \url{https://github.com/patcg-individual-drafts/topics/blob/main/taxonomy_v2.md}\\

\noindent Note:
\begin{enumerate}
    \item Do differentiate between a domain and a topic. A topic of the sentence is the main idea of the sentence. Where this sentence belongs is the domain. In this task we are classifying topics.
    \begin{enumerate}
        \item “Aiden was once a warrior who placed complete faith in his own abilities.” - this belongs to literature/creative writing domain, but the topic of the sentence is \topic{Conversation \& Opinion}. 
        \item “i keep telling you to leave me alone, this forum is not the right place for hate” - the domain is media, but the topic is \topic{Conversation \& Opinion}.
    \end{enumerate}
\item If needed, please do a quick search of the sentence to identify the topic. Do limit the search to a quick scan of search results no longer than 30 sec.
\item If a sentence fits more than one topic equally and you cannot decide between the two, then select a primary and a secondary topic. Add a comment if needed to explain. Try not to abuse this option and always try to choose one.
\end{enumerate}

\end{document}